\documentclass[runningheads]{llncs}

\usepackage{eccv}

\usepackage{caption}
\usepackage{eccvabbrv}

\usepackage{graphicx}
\usepackage{booktabs}

\usepackage{multirow} 
\usepackage{booktabs}   % optional, for nicer rules (\toprule, \midrule, \bottomrule)
\usepackage{tabularx}
\newcolumntype{Y}{>{\raggedright\arraybackslash}X}

\usepackage{makecell}
\usepackage{amsmath}

\usepackage[accsupp]{axessibility}  % Improves PDF readability for those with disabilities.

\usepackage{hyperref}

\usepackage{orcidlink}

\DeclareMathOperator{\argmin}{argmin}
\begin{document}

% \title{3D-Consistent Multi-View Editing by Diffusion Guidance} 
\title{3D-Consistent Multi-View Editing by Correspondence Guidance} 

\titlerunning{3D-Consistent Multi-View Editing by Correspondence Guidance}

\author{Josef Bengtson\inst{1} \and
David Nilsson\inst{1} \and
Dong In Lee\inst{2} \and Yaroslava Lochman \inst{1} \and Fredrik Kahl\inst{1}}
\authorrunning{J.~Bengtson et al.}
\institute{Chalmers University of Technology \email{\{bjosef,david.nilsson,lochman,fredrik.kahl\}@chalmers.se}\and Korea University\\
\email{dilee99@korea.ac.kr}
}

\maketitle

\begin{abstract}
Recent advancements in diffusion and flow models have great\-ly improved text-based image editing, yet methods that edit images independently often produce geometrically and photometrically inconsistent results across different views of the same scene. Such inconsistencies are particularly problematic for editing of 3D representations such as NeRFs or Gaussian splat models. We propose a training-free guidance framework that enforces multi-view consistency during the image editing process. The key idea is that corresponding points should look similar after editing. To achieve this, we introduce a consistency loss that guides the denoising process toward coherent edits. The framework  is flexible and can be combined with widely varying image editing methods, supporting both dense and sparse multi-view editing setups. Experimental results show that our approach significantly improves 3D consistency compared to existing multi-view editing methods. We also show that this increased consistency enables high-quality Gaussian splat editing with sharp details and strong fidelity to user-specified text prompts. Please refer to our project page for video results: \href{https://3d-consistent-editing.github.io/}{https://3d-consistent-editing.github.io/}
  \keywords{Multi-View Editing \and Diffusion models \and 3D Consistency}
\end{abstract}

\section{Introduction}
\label{sec:intro}
Text-based image editing has become increasingly powerful with generative models, allowing modification of an image with a prompt such as ``turn the person into a clown'' or ``make it foggy''. While these methods show impressive results when editing single images, using them to edit multiple images of the same scene, or to edit 3D models, in a consistent manner remains difficult.

\begin{figure}[t]
    \centering
    \includegraphics[width=\linewidth]{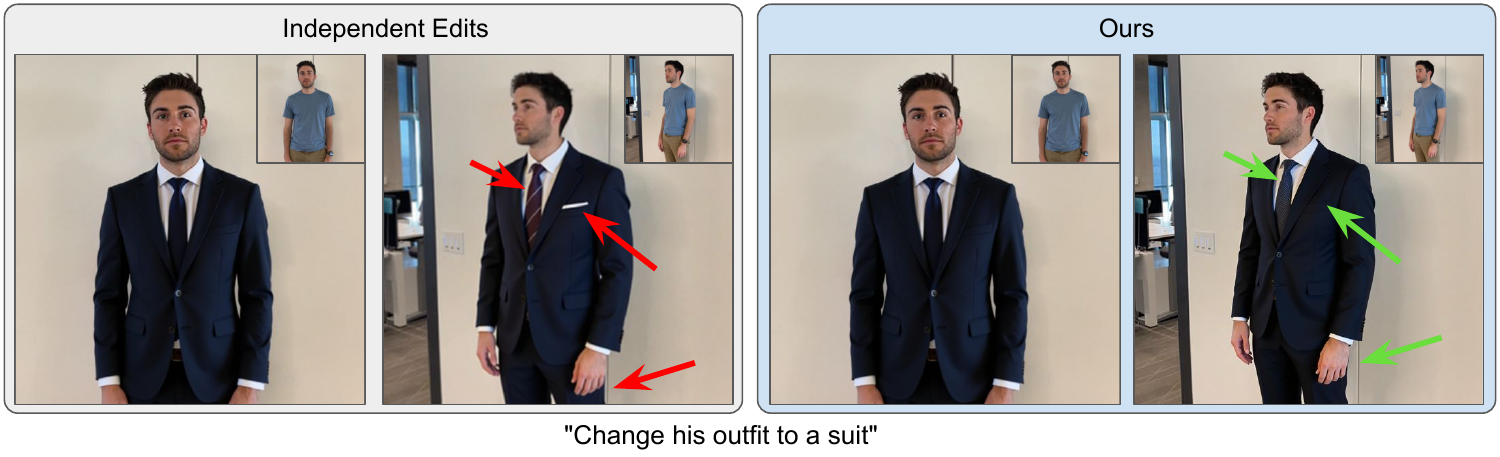}
    \includegraphics[width=\linewidth]{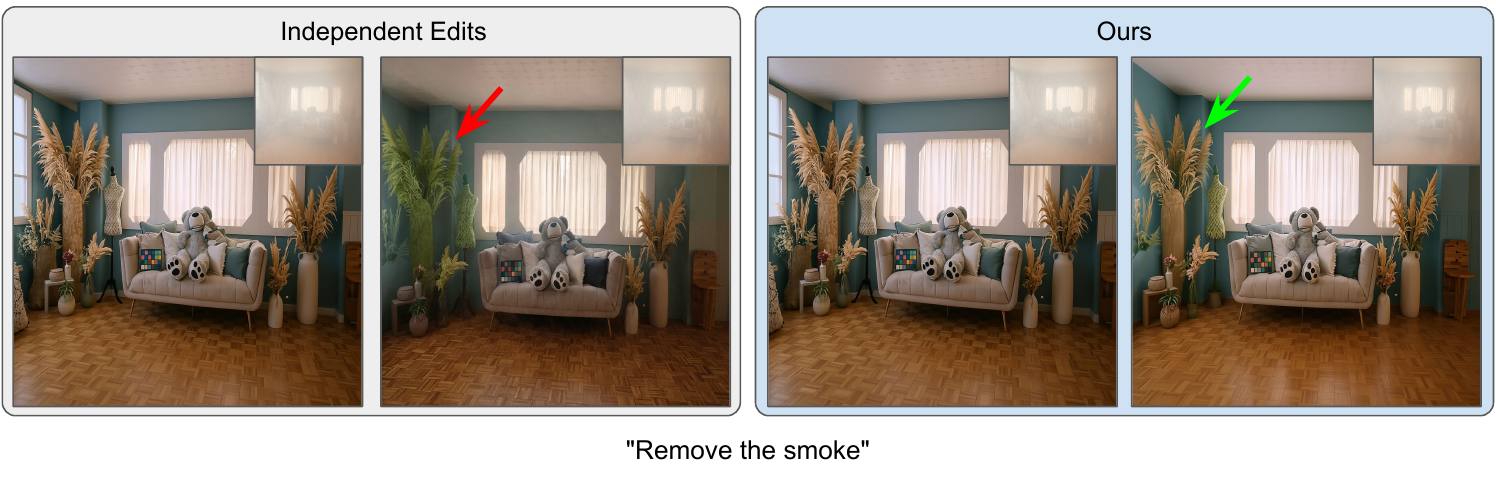}
    \caption{
Image editing methods applied independently to multi-view images often produce inconsistent edits across views, e.g.\ in the tie pattern and the location and color of the decorations (red arrows). Our method improves multi-view consistency (green arrows) by guiding the denoising process so that corresponding points are edited similarly. Object poses also better match the original images.
    }
    \label{fig:intro_fig}
\end{figure}

Given recent work on photorealistic rendering of 3D models using NeRFs \cite{mildenhall2021nerf} or Gaussian splatting \cite{kerbl20233d}, a natural question is whether we can edit such 3D models using image editing. If the training views that are used to train 3D models are edited independently, then there is an issue that different images are altered with different characteristics. For example, changing a person's outfit to a suit as in Fig. \ref{fig:intro_fig} can look realistic for each image, but different characteristics, such as the tie pattern and handkerchief visibility, are inconsistent across the different views. For 3D editing, this implies that the edited training views do not depict the same 3D scene, making the task difficult. This motivates modifying the image editing to be multi-view consistent. Early work on 3D editing such as Instruct-NeRF2NeRF \cite{haque2023instruct} addresses the multi-view issue by alternating between editing all images and updating the 3D model until convergence. Recent works such as EditSplat \cite{lee2025editsplat} and DGE~\cite{chen2024dge} modify image generation using multi-view constraints or attention to improve consistency, and directly update the 3D model, bypassing the costly iterative regeneration of edited images.

We present a method to directly guide the denoising process of diffusion and flow-matching image editing methods so that the edited images are multi-view consistent, as shown in Fig. \ref{fig:intro_fig}. Our key idea is that corresponding points across images should look similar after editing. Based on this idea, we introduce a simple guidance loss that measures multi-view consistency. One way to approach this is by matching points between \textit{unedited} images and forcing these points to be changed in a similar way. This approach does not allow for larger geometry changes. While the primary focus of our work is on \textit{near-rigid} edits, we also find that using the same loss with matches between \textit{edited} images improves multi-view consistency in \textit{non-rigid} edits, i.e., edits with larger geometry changes. During image generation, we enforce the consistency by using training-free guidance which modifies the denoising process so that it is guided towards samples where our consistency loss has a low value. 

As an application of our multi-view consistent image editing, we show how to edit 3D Gaussian splat models. We simply resume training based on the edited images which are now sufficiently multi-view consistent to get a Gaussian splat model with sharp and realistic details. Our method also enables consistent editing of sparse views, and it works with different image editing methods. 

In summary, our main contributions are: 
\begin{itemize}
    \item We introduce a flexible training-free method able to guide widely varying image editing models to generate multi-view consistent images.
    \item Our method is based on the key idea that corresponding points should look similar in the edited images. This is enforced by optimizing the proposed consistency loss on corresponding points across views.
    \item We show that the generated images can be used to directly refine a Gaussian splat model and accurately reconstruct fine details.
    \item We compare our method to recent work and show improved multi-view consistency of the generated images prior to Gaussian splat refinement, and we obtain edited Gaussian splat models
    faithful to the given text prompts with clear details, demonstrated through extensive video comparisons.
\end{itemize}

\begin{figure}[t]
    \centering
    \includegraphics[width=0.97\linewidth]{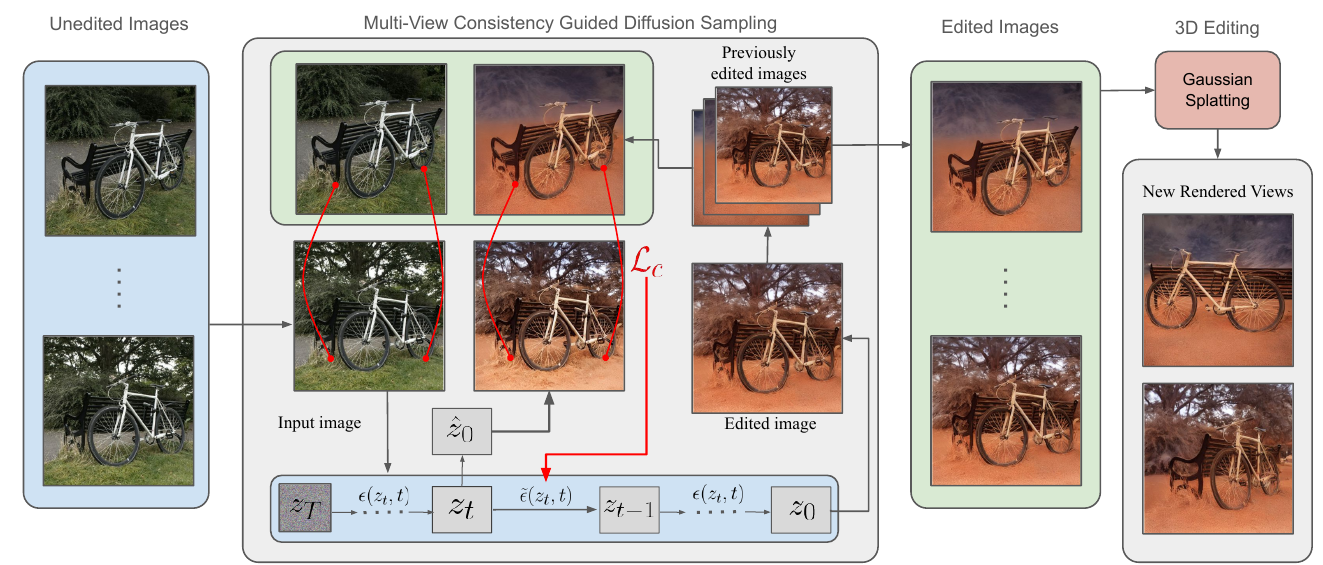}
    \caption{Overview of our method. Given input images, each view is edited sequentially by guiding the denoising process using previously edited images. The guidance assumes corresponding points should look similar in the edited images. During denoising, the noise estimate $\epsilon(z_t,t)$ is modified according to a consistency loss $\mathcal{L}_c$, producing multi-view consistent edits. Edited images are then used to update a Gaussian splat model.}
    \label{fig:system_figure}
\end{figure}

\section{Related Work}
\label{sec:related_work}

\subsection{2D Image Editing}
The development of powerful generative models \cite{rombach2022high,goodfellow2014generative} has led to large advancements in image editing, initially mainly through Generative Adversarial Networks (GANs) \cite{goodfellow2014generative,wang2018high,zhu2017toward,CycleGAN2017} and more recently through models based on diffusion \cite{qwen,palette,diffuseit,kawar2023imagic} and flow matching \cite{flux,flowlatentediting,kulikov2025flowedit}. InstructPix2Pix \cite{brooks2022instructpix2pix} finetunes Stable Diffusion \cite{rombach2022high} using image pairs generated by prompt-to-prompt (P2P) \cite{hertz2022prompt} following instructions generated by GPT-3 \cite{brown2020language}, enabling instruction-based editing. However, its limiting factor is the long editing time due to the large number of denoising steps. Subsequent methods therefore employ few-step \cite{turboedit,infedit,cora,instantedit,mei2024conditional} or one-step \cite{parmar2024one, swiftedit,lai2025instantportrait,zhao2024fastdrag} denoising. Our work focuses on how to make such methods multi-view consistent, enabling direct 3D editing. 

\subsection{3D Scene Editing}

One approach to 3D scene editing is to utilize paired 3D data to directly edit a 3D representation \cite{xia2025scalableconsistent3dediting,ye2025nano3dtrainingfreeapproachefficient,li2025voxhammer,ava-nvs}, but such data is expensive to acquire, limiting training to single objects and synthetic scenes. 
Another popular approach leverages existing image editing methods for 3D editing, but these 2D methods provide no consistency guarantees across views. Some works use score distillation sampling (SDS)~\cite{poole2023dreamfusion} to update the 3D representation \cite{kamata2023instruct, zhuang2023dreameditor, sella2023voxe,focaldreamer}. Instruct-NeRF2NeRF \cite{haque2023instruct} uses consistency from a 3D representation to achieve consistent edits by iteratively editing views and updating the 3D representation. This idea, known as iterative dataset update, is widely used in follow-up works \cite{chen2023gaussianeditor,vcedit,mirzaei2023watchyoursteps,ProteusNeRF,dong2023vica}. 

The iterative process is compute-intense and several approaches make the editing process more consistent to reduce dataset update iterations. Improvements range from incorporating correspondence regularization into the denoising process \cite{song2023efficient,edicho}, scene-specific finetuning \cite{C3Editor}, utilizing geometric information from the 3D representation or monocular depth estimation to guide the editing \cite{diffusion3dfield,lee2025editsplat,fujiwara,gaussctrl2024} and allowing the attention process in the editing method to consolidate information across views \cite{intergsedit,chen2024dge,gaussctrl2024,vcedit,fujiwara}. Several of these recent approaches \cite{chen2024dge,vcedit} rely on iterative refinement on top of multi-view consistent editing. In contrast, our method provides further consistency improvements and does not require updating the images. Our method can also handle sparse view editing, while several baseline methods are limited to dense views, since they either require geometric information from an existing 3D representation \cite{lee2025editsplat,gaussctrl2024} or require rendering a smooth camera trajectory, inspired by video editing methods \cite{chen2024dge}.

\subsection{Training-Free Guidance}
There exist different approaches for adapting a diffusion or flow model to a specific task. One approach is to fine-tune the model \cite{hu2022lora,zhang2023adding}, which requires significant amounts of training data and compute. A contrasting approach is to guide the sampling process of an existing model without additional training. This does not require training data but instead a loss function to minimize. In training-free guidance methods, each step in the denoising process uses the current noisy sample to predict the clean sample and compute a per-step correction \cite{bansal2024universal,FreeDoM,TFG,he2024manifold,patel2024flowchef,sun2024rectifid}. Another approach is to optimize the initial noise for the denoising process \cite{Samuel2023NAO,Samuel2023SeedSelect,bengtson2025geometric,SONIC,ronai2025flowopt}, allowing direct optimization guided by the loss function, computed via gradients through all denoising steps. Any loss function can be used as guidance, e.g., from an image classifier or segmentation model. There are also methods that use losses for 3D consistency for single-image novel view synthesis \cite{bengtson2025geometric,you2025nvssolver}, or solve image inpainting based on geometric correlation to a reference image \cite{liu2025corrfill}. In this work, we apply training-free guidance to image editing methods to improve their multi-view consistency. 

\section{Method}
\label{sec:method}

We start with an overview, followed by an introduction to our consistency loss. We then show how to optimize a set of images for multi-view consistent edits. Finally, we describe how these images can be used to edit a Gaussian splat model.

\subsection{Overview}
We show an overview of our method in Fig. \ref{fig:system_figure}. We propose a method to edit a set of images in a 3D consistent way. Our method can be seen as an extension of 2D image editing methods like \cite{brooks2022instructpix2pix, parmar2024one, flux}, where images are edited based on a text prompt. 
Applying 2D editing methods on each image independently results in inconsistent edits. Our method guides the denoising process such that the edits are consistent across the views. We use the fact that corresponding points in the edited images should have similar perceptual features. Our edited images can be used directly to edit a Gaussian splat model of the scene.

\subsection{Consistency Loss on Correspondences}
Our main optimization criterion for 3D-consistent editing is based on the assumption of similar appearances for corresponding points. In particular, (1) if an edit preserves geometry, corresponding points in the unedited images $I_1$ and $I_2$ should be edited in a similar way in the edited images $I'_1$ and $I'_2$, and (2) if an edit changes geometry, the corresponding points in $I'_1$ and $I'_2$ should have similar features. We define the consistency loss $\mathcal{L}_c$ as
\begin{equation}\label{eq:consistency_loss}
    \mathcal{L}_c = \sum_{(x,y)\in \mathcal{M}}\left(\|I'_1(x) - I'_2(y)\|_1 +\\ \lambda f_\text{LPIPS}(I'_1(x), I'_2(y))\right),
\end{equation}
where $f_\text{LPIPS}$ is the perceptual loss \cite{zhang2018perceptual} (with $\lambda$ set to $2$) applied to patches centered around the matches, and $\mathcal{M}$ is the set of corresponding points. To obtain corresponding points, we match images from either the unedited pair $\left(I_1, I_2\right)$ or the edited pair $\left(I'_1, I'_2\right)$. Using matches between unedited images ensures that the geometric details are preserved, which is beneficial in rigid and near-rigid editing, such as changing textures, materials, colors and weather conditions. On the other hand, for non-rigid editing (like changing objects or adding new objects), using matches between edited images additionally forces consistency in the areas where geometry is changed. For this to work, it is important to use a robust matcher that has an understanding of global context and semantics. We use the robust dense matcher RoMa \cite{edstedt2024roma}. More details on hyperparameters can be found in appendix \cref{sec:suppl_mw_edit}.

\subsection{Consistency Guided Image Editing}
Here we describe how to use our consistency loss to adapt a denoising process to generate a set of multi-view consistent edits.
Our method is based on pre-trained image editing models, and we employ different training-free methods to guide the denoising process, as is detailed below.

To use our consistency loss with the diffusion model InstructPix2Pix \cite{brooks2022instructpix2pix}, we employ universal guidance \cite{bansal2024universal}, where the sampling is steered towards low values of a loss function $\mathcal{L}(z)$, which in our case is the consistency w.r.t.\ previous edits. The noise estimation $\epsilon(z_t, t)$ is modified to include a correction based on gradients of the loss function $\mathcal{L}$ to obtain a new noise estimate $\tilde{\epsilon}(z_t, t)$ that guides the latent to low values of the loss. The modified noise is computed as
\begin{equation}
    \tilde{\epsilon}(z_t, t) = \epsilon(z_t, t) + \lambda_{t}\nabla_{z_t} \mathcal{L}(\hat{z}_0(z_t))
\end{equation}
where $\hat{z}_0(z_t)$ is a one-step prediction of the denoised latent $\hat{z}_0(z_t)=\frac{1}{\sqrt{\alpha_t}}(z_t - \sqrt{1-\alpha_t}\epsilon(z_t, t))$, and $\lambda_t$ is the weight of the guidance. For our application, the best results are obtained using $\lambda_t = \lambda \mathbf{1}(t < N_g)$, so that we activate the guidance only for the last $N_g=700$ steps of the denoising process, after first running the denoising process without any guidance. We also use backward guidance \cite{bansal2024universal}, where we optimize a correction $\Delta z_0 = \argmin_\Delta \mathcal{L}(\hat{z}_0 + \Delta)$ for $N_b$ gradient descent steps, and update the noise prediction as $\tilde{\epsilon}'(z_t, t) = \tilde{\epsilon}(z_t, t) - \sqrt{\alpha_t / (1 - \alpha_t)}\Delta z_0$.

We can also use our consistency loss with the one-step method pix2pix-Turbo \cite{parmar2024one}, where the denoising process is reduced to a single step. Such a model can be formulated as $x=f(z)$, where $z$ is the starting noise and $x$ is the generated image, and we optimize $\mathcal{L}(f(z))$ with respect to $z$. This optimization resembles SeedSelect \cite{Samuel2023SeedSelect}, where the starting noise is optimized to constrain the sampling process. This is similar to what has previously been used to improve geometric consistency of diffusion models for single-image novel view synthesis \cite{bengtson2025geometric}. 

Our consistency loss can also be used to guide flow matching models. We utilize an efficient optimization strategy proposed by \cite{SONIC} that uses a linear approximation of the denoising trajectories to reformulate the optimization objective as $\mathcal{L}([f(z)-z]_{\text{sg}}+z)$, where $f(z)$ is the output of the flow matching process, $z$ is the initial noise and $[\cdot]_{\text{sg}}$ is the stop-gradient operation. The difference $f(z)-z$ is therefore seen as constant throughout the denoising process. This removes the expensive unrolling related to computing gradients through the denoising process with many steps and enables efficient guiding of flow matching models with our consistency loss. More details regarding this  can be found in appendix \cref{sec:suppl_mw_edit}.

\paragraph{Image Ordering} An important consideration when editing multiple images of the same scene is how to select which previously edited image to use when computing the consistency loss $\mathcal{L}_c$. We found that editing one image at a time works well in practice, and when we generate a new image, we use the matches between that image and two of the previously edited images. We select the images with the most matching points to the currently edited image. We found that there is no performance improvement when using more than two images (see Sec. \ref{sec:ablations}).
\begin{figure}[t]
    \centering
    \textbf{Edited Multi-View Images with InstructPix2Pix}\\
    \includegraphics[width=\textwidth,trim={0 5px 0 0},clip]{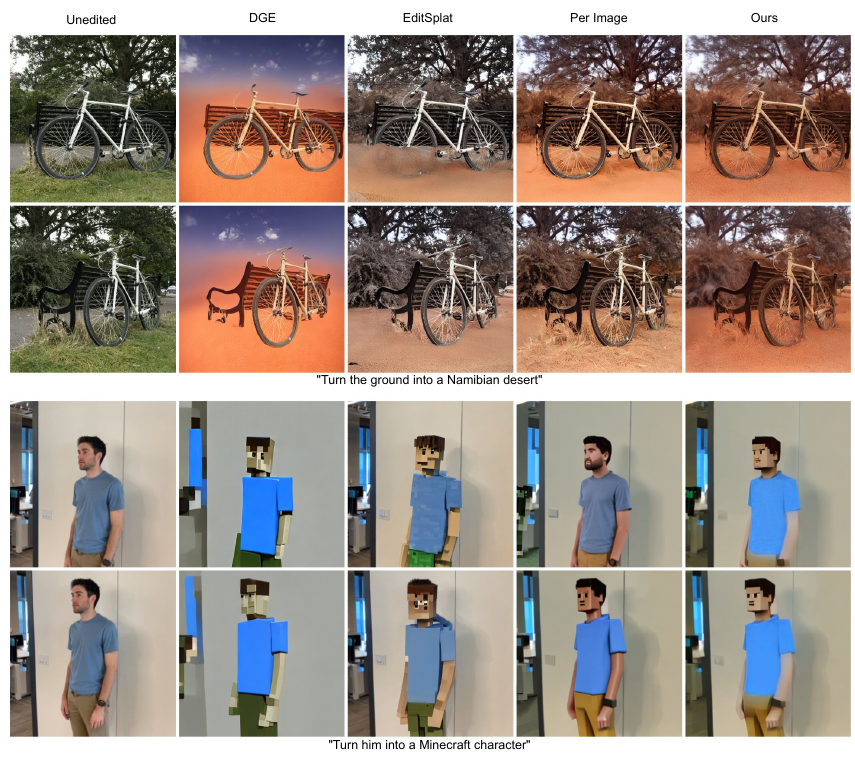}
    \caption{Qualitative comparison using InstructPix2Pix. EditSplat and DGE edit images more drastically. Our method preserves the scene better, as can be seen e.g., in the texture of the grass next to the bench, and in the shape of the face, the gaze and the preserved watch. Our method also produces more consistent edits between different views, as can be seen on the bicycle wheels, or on the face or arms of the person.
    }
    \label{fig:image_consistency}
\end{figure}

\subsection{3D Gaussian Splat Editing}\label{sec:editing_3DGS}
For editing 3D Gaussian splat models, we start with a trained Gaussian splat model obtained from the unedited views and then resume training using the consistently edited images for 20 epochs. It is also possible to use only a sparse set of (e.g., 3-4) images. In that case, a multi-view diffusion method like ViewCrafter \cite{yu2024viewcrafter} can generate novel views of the edited scene, using poses interpolated between the edited views. The edited images and generated views can then be used to train a Gaussian splat model representing the edited scene. Additional details are in appendix \cref{sec:suppl_GS_edits}.

\begin{figure*}[t]
    \centering
    \textbf{Renderings from Edited 3DGS Models}\\
    \includegraphics[width=0.97\textwidth,trim={0 5px 0 0px},clip]{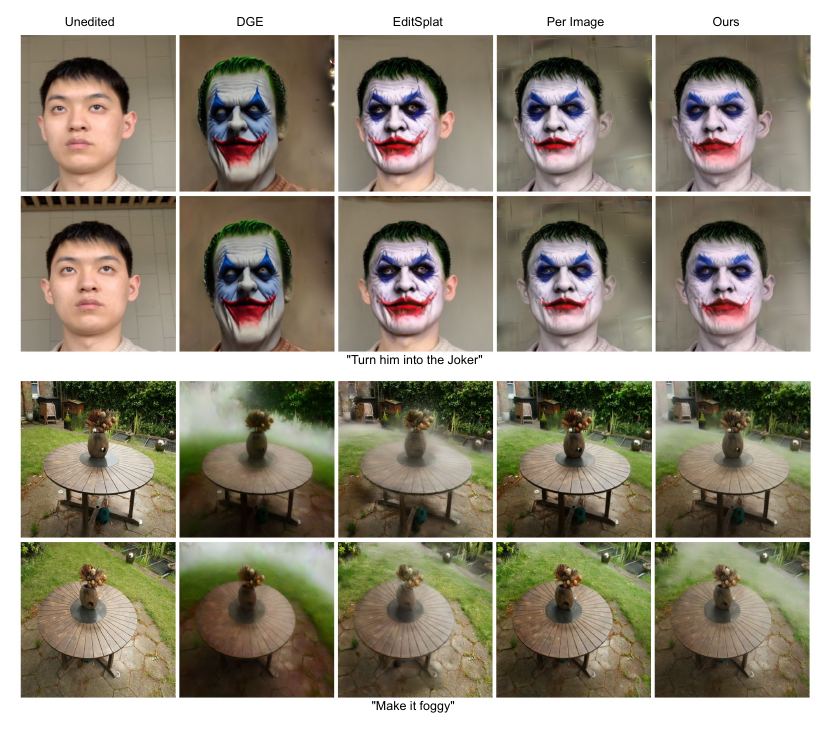}
     \caption{Renderings from edited 3D Gaussian splat models. For the top scene, per-image edits sometimes cause blurriness, as seen, e.g., in the ears which are sharper for ours. EditSplat uses a segmentation mask, leading to the edit being localized only on the face and hair. For DGE, the fidelity to the text prompt is high, but the face geometry has drastically changed. For the bottom scene, our method gives a clearer edit than all other methods, as seen by more visible fog and sharp details preserved on the objects.}
    \label{fig:images_full_pipeline}
\end{figure*}

\section{Experiments}
\label{sec:experiments}
We evaluate our method in two main ways, namely for multi-view consistent image editing and 3D editing of Gaussian splat models. We also provide ablation studies and results when editing just a sparse set of views. In addition to the results presented here we on the project page (\href{https://3d-consistent-editing.github.io/}{https://3d-consistent-editing.github.io/}) provide video results showing comparisons against the baseline methods for all the scenes.

\subsection{Setup}

\paragraph{Implementation Details}
To evaluate our model, we use the same 8 test scenes as in EditSplat \cite{lee2025editsplat}, including real-world scenes from IN2N \cite{haque2023instruct}, Mip-NeRF360 \cite{barron2022mipnerf360}, and BlendedMVS \cite{blendedmvs}. For validation, we use 2 scenes from  Mip-NeRF360 \cite{barron2022mipnerf360}, 1 scene from IN2N \cite{haque2023instruct}, and 1 self-captured scene containing images of a person's head, comparable to the IN2N ``Face'' scene.  For the test scenes we use a total of 21 different edit prompts and for the validation scenes we use a total of 7 different edit prompts. The exact prompts are provided in appendix \cref{sec:suppl_scene_prompt_pairs}. The scenes are of varying size, containing 65-350 images per scene. All experiments are done on a single A100 GPU. The complete editing process for the ``Face'' scene from IN2N takes about 22 minutes for our method when using InstructPix2Pix and about 17 minutes when using pix2pix-Turbo.

\paragraph{Baselines}
Since our main goal is consistent multi-view editing, we wanted to compare with other methods that also aim to achieve this; therefore, we chose EditSplat \cite{lee2025editsplat} and DGE \cite{chen2024dge} since they are state-of-the-art methods for 3D editing that claim multi-view consistency during the 2D editing process.
Another recent method that claims improved multi-view consistency is InterGSEdit \cite{intergsedit}, but it does not have code available. Both EditSplat and DGE use the image editing model InstructPix2Pix \cite{brooks2022instructpix2pix}, so for fairness we also evaluate our method using it. We also test our method on the one-step model pix2pix-Turbo \cite{parmar2024one} and flow-matching based model FLUX.1 \cite{flux}. InstructPix2Pix and pix2pix-Turbo have limited ability in changing geometry, therefore we use near-rigid edits for these methods and optimize the consistency loss based on matches between unedited images. When we evaluate image consistency, we extract the images \emph{prior} to updating the Gaussian splats, which, for EditSplat, is the output from their Multi-view Fusion Guidance (MFG), and for DGE, is the output after one iteration of their Multi-View Consistent Editing. We also compute metrics for the unedited images to get upper bounds on the consistency. As a baseline, we also edit per image without any consideration of multi-view consistency.

\paragraph{Metrics}
There are two aspects of the generated images we evaluate, namely  multi-view consistency and fidelity to the text prompt. For multi-view consistency, we evaluate MEt3R \cite{asim2025met3r} that compares DINO \cite{Caron2021EmergingPI} embeddings of matches obtained via Dust3R \cite{wang2024dust3r}. Additionally, we evaluate multi-view consistency by computing the rendering metrics PSNR, SSIM and LPIPS by training a Gaussian splat model \cite{kerbl20233d} from scratch using only the edited images. The edited images are split into training and test sets, and the metrics are computed on the edited images set aside as test images. Inconsistent views yield blurrier renderings and thus lower reconstruction metrics. We also evaluate how well the edited images match the given text prompts by comparing CLIP \cite{radford2021learning} embeddings of images and text.
Given a reference image $I$, an edited image $I'$ and text descriptions\footnote{Similar to previous work \cite{lee2025editsplat}, we have three prompts per scene: a description of the unedited image, a description of the edited image and the edit prompt, for instance ``a photo of a park'', ``a photo of a Namibian desert'', and ``Turn the ground into a Namibian desert'', respectively.} of the unedited and edited images $T$ and $T'$, respectively, CLIPdir measures the cosine similarity between $\text{CLIP}(I')-\text{CLIP}(I)$ and $\text{CLIP}(T')-\text{CLIP}(T)$, CLIPsim is the cosine similarity between $\text{CLIP}(I')$ and $\text{CLIP}(T')$. Both measure how well the edited image aligns with the desired edit. Finally, CLIPimage is the cosine similarity between $\text{CLIP}(I)$ and $\text{CLIP}(I')$, capturing how well the edited image preserves the content of the input image. There is an inherent trade-off between CLIPimage and CLIPdir/CLIPsim: pushing the edited image to align more closely with the prompt can reduce its similarity to the input image, causing it to lose input details that should ideally be preserved.

\subsection{Image Consistency Evaluation}

Here we evaluate the multi-view consistency of the edits.
For fair comparison, we extract the images \emph{prior} to training Gaussian splats, in contrast to the results of Gaussian splat editing in Sec. \ref{sec:full_methods}. The image editing consistency is presented in Table \ref{tab:image_editing}. For the consistency metrics, we can see that our method obtains the best values of all methods, showing that our edited images are more multi-view consistent than both the per-image baseline and the two methods we compare to.
\begin{table*}[t]
\centering
\begin{tabular}{lcccccccc}
\hline
\textbf{Method}    & MEt3R$\downarrow$ & PSNR$\uparrow$ & SSIM$\uparrow$ & LPIPS$\downarrow$ & CLIPdir$\uparrow$ & CLIPsim$\uparrow$ & CLIPimage$\uparrow$ \\ \hline
Unedited           & 0.183              & 27.36           & 0.815           & 0.190              & -                  & 0.199              & 1.0                    \\ \hline \hline
 \multicolumn{8}{c}{Based on InstructPix2Pix} \\ \hline \hline
EditSplat          & 0.329              & 20.20           & 0.641           & 0.389              & 0.159              & 0.252              & 0.768                    \\
DGE                & 0.224              & 21.58           & 0.705           & 0.256              & \textbf{0.173}     & \textbf{0.257}     &0.757                    \\ \hline
Per image      & 0.243              & 21.19           & 0.679           & 0.271              & 0.153              & 0.249              & 0.813                    \\
Ours          & \textbf{0.212}     & \textbf{23.46}  & \textbf{0.716}  & \textbf{0.247}     & 0.152              & 0.249              &\textbf{ 0.817 }                  \\ \hline \hline
\multicolumn{8}{c}{Based on pix2pix-Turbo} \\ \hline \hline
Per image & 0.291          & 18.29           & 0.564           & 0.429              & \textbf{0.124}              & \textbf{0.254}              & 0.766                    \\
Ours & \textbf{0.226}              & \textbf{24.73}           & \textbf{0.696 }          &\textbf{ 0.339  }            & 0.116              & 0.252              & \textbf{0.768}                   \\ \hline
\end{tabular}
\vspace{1em}
\caption{Evaluation for multi-view consistent editing. Our method significantly improves the consistency with both InstructPix2Pix \cite{brooks2022instructpix2pix} and pix2pix-Turbo \cite{parmar2024one}. For DGE and EditSplat we note decreased consistency metrics, but also that CLIPsim and CLIPdir are improved and CLIPimage decreases which indicates larger edits to the images than our method and the per image baseline.}
\label{tab:image_editing}
\end{table*}
We note that for both EditSplat and DGE, CLIPimage is lower than for InstructPix2Pix per image, indicating that the multi-view edited images aggregate edits that are stronger than the baseline edit method.
For the text metrics there is no clear overall improvement of our method compared to the per-image edits, which is expected since we optimize for multi-view consistency only. 
We show qualitative results in Fig. \ref{fig:image_consistency}. Comparing our method to the per-image edits, we see that our edits have significantly improved multi-view consistency. While EditSplat and DGE are more consistent than the per-image edits, some inconsistencies still remain, e.g., on the bicycle wheels or on the face of the person. 

\subsection{Gaussian Splat Editing}\label{sec:full_methods}

As an application of our multi-view consistent image editing, we show how to use the edited images to update 3D Gaussian splats. We show the results in Table \ref{tab:full_pipeline}. We note that CLIPsim and CLIPdir are comparable for our method and EditSplat, both achieving similar values as the per-image edits. DGE shows improved values for CLIPdir and CLIPsim but worse on CLIPimage which indicates stronger edits that are less similar to the unedited images. Note that CLIPsim and CLIPdir measure how well the images fit the text prompt, and only loosely measure other relevant aspects such as sharpness of the images. The increased MEt3R score for DGE shows that the renderings for this method are less consistent than the others.

We show qualitative examples in Fig. \ref{fig:images_full_pipeline}. We notice that updating Gaussian splats with independently edited images can cause blurry results compared to using multi-view consistent editing. This can be seen, e.g., at the ear of the person or the fog over the grass which is only present in a few of the independently edited images and only weakly seen in the Gaussian splat model. 
\begin{table}[t]
\centering
\begin{tabular}{lccccc}
\toprule
 & CLIPdir$\uparrow$ & CLIPsim$\uparrow$ & CLIPimage$\uparrow$ & MEt3R$\downarrow$ \\ \hline \hline
 \multicolumn{5}{c}{Based on InstructPix2Pix} \\ \hline \hline
EditSplat & 0.123 & 0.238 & 0.832 & \textbf{0.215} \\
DGE & \textbf{0.146} & \textbf{0.242} & 0.752 & 0.242 \\ \hline
Per image & 0.126 & 0.239 & \textbf{0.833} & 0.216 \\
Ours & 0.121 & 0.237 & 0.830 & \textbf{0.215} \\  \hline \hline
 \multicolumn{5}{c}{Based on pix2pix-Turbo} \\ \hline \hline
 Per image & \textbf{0.067} & \textbf{0.224} & \textbf{0.845} & \textbf{0.210} \\ 
Ours& \textbf{0.067 }& 0.223 & 0.823 & \textbf{0.210} \\ \bottomrule
\end{tabular}
\vspace{1em}
\caption{Evaluation of 3D Gaussian splat editing. The renderings from the edited Gaussian splat models are similarly accurate with respect to the text prompts as when using the per-image edits. We note that DGE gets the highest values of CLIPdir and CLIPsim but the lowest for CLIPimage, indicating more substantial edits of the scene. The Met3R score here is evaluated on the rendered views, and measures how consistent the renders are across different viewing directions.}
\label{tab:full_pipeline}
\end{table}
These cases are handled better by our consistency guided denoising. We also note that DGE can result in edits where the scene content is modified significantly, e.g., the shape of the face is modified so that it no longer resembles the person in the input image. We provide several more examples in the videos on the project page, where we see that our method can produce clear edits with preserved details.

\begin{figure*}[t]
    \centering
    \textbf{Edited Images and Renderings with pix2pix-Turbo}\\
    \includegraphics[width=\textwidth,trim={0 5px 0 0},clip]{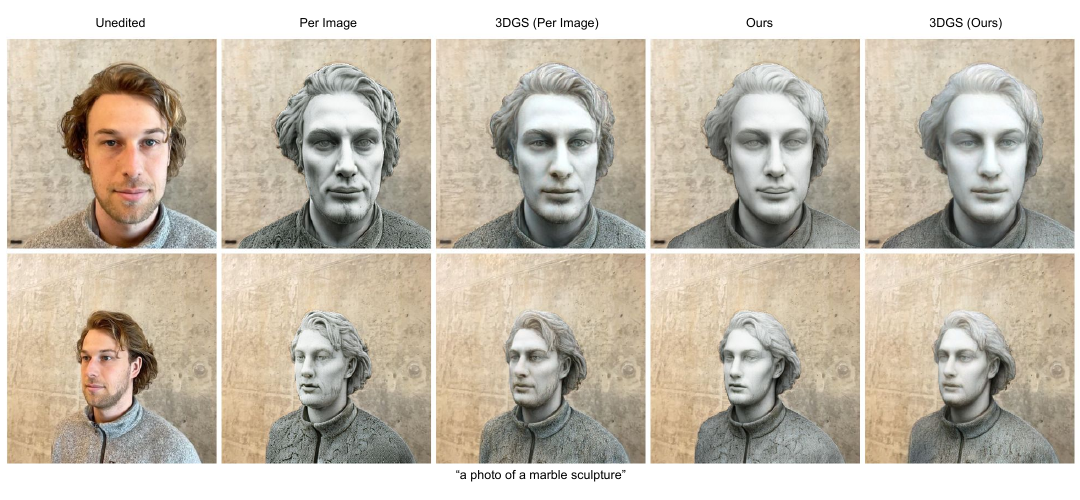}
    \caption{Multi-view editing using the image editing method pix2pix-Turbo. Per-image edits can be inconsistent and there is a loss of detail when editing a 3D Gaussian splat model using these inconsistent images. In contrast, our edits are more consistent and the details are more accurately recovered by the Gaussian splat renderings.}
    \label{fig:pix2pixturbo}
\end{figure*}

To show that our method is not limited to InstructPix2Pix, we also use it with pix2pix-Turbo \cite{parmar2024one} and present results in Fig. \ref{fig:pix2pixturbo}. The per-image edits are inconsistent in several places, e.g., colors around the eyes. Edits using our correspondence guidance are significantly more consistent which can also be seen for the Gaussian splat renderings.

\begin{table*}[t]
\centering\begin{tabular}{lccccc}
\toprule
\textbf{} & PSNR$\uparrow$ & SSIM$\uparrow$ & LPIPS$\downarrow$ & \multicolumn{1}{l}{CLIPsim$\uparrow$} & MEt3R$\downarrow$ \\
\midrule
Per Image & 22.20          & 0.687      & 0.290         & 0.259                               & 0.404            \\
Ours      & \textbf{24.49 }         & \textbf{0.725 }    & \textbf{0.281 }      &\textbf{ 0.263  }                             &\textbf{0.364 } \\
\bottomrule
\end{tabular}
\vspace{1em}
\caption{Evaluation of final renderings for the sparse setting where 3-4 views are given as input. Our method combined with pix2pix-Turbo \cite{parmar2024one} is used to edit the images, and the multi-view diffusion model ViewCrafter \cite{yu2024viewcrafter} is used to generate additional edited views that are then used to train a 3DGS representation. We evaluate on four scenes (a total of 6 prompts). The metrics are computed w.r.t. a few of the generated images used as a test set. The consistency is significantly higher for our method compared to using independent edits per image.}
\label{tab:sparse}
\end{table*}
\subsection{Sparse Editing}
\begin{figure}[h]
    \centering
    \textbf{Sparse Edits and Novel View Renderings}\\
    \includegraphics[width=\linewidth]{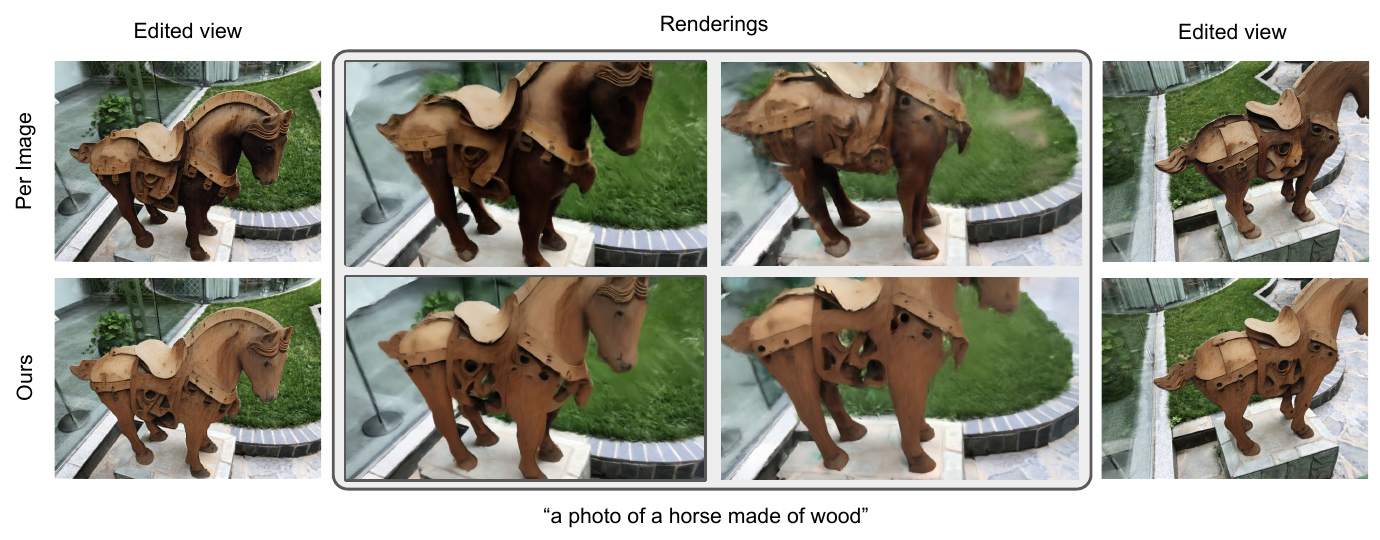}
    \caption{Example of editing a few sparse views and using the multi-view diffusion model ViewCrafter \cite{yu2024viewcrafter} to interpolate. We note that inconsistencies in the edited views lead to blurry renderings and view inconsistencies in the interpolated images. When using our edited images the interpolated views are both more consistent and contain sharper details.}
    \label{fig:sparse}
\end{figure}

As another application of our multi-view consistent editing, we also investigate editing just a sparse set of 3-4 images and interpolating with the multi-view diffusion model ViewCrafter \cite{yu2024viewcrafter} to generate additional edited views that can then be used to train a Gaussian splat model representing the edited scene (see Sec. \ref{sec:editing_3DGS}), instead of editing 40-125 images. Similarly to the image consistency evaluation, we use a few of the views as test views to evaluate the Gaussian splat model. We show the results in Fig. \ref{fig:sparse} and Table \ref{tab:sparse}. We note that our sparsely edited views are more consistent, which makes the rendering significantly more consistent and sharp.

\subsection{Flexibility of Consistency Guidance}\label{sec:felxibility_flux}
To showcase the flexibility of our consistency-based guidance approach, we present the qualitative results using the recently published method FLUX.1 \cite{flux}. 

\begin{figure}[h]
    \centering
    \textbf{Edited Multi-View Images with FLUX.1}
   
    \includegraphics[height=190px,trim={0 0 0 0},clip]{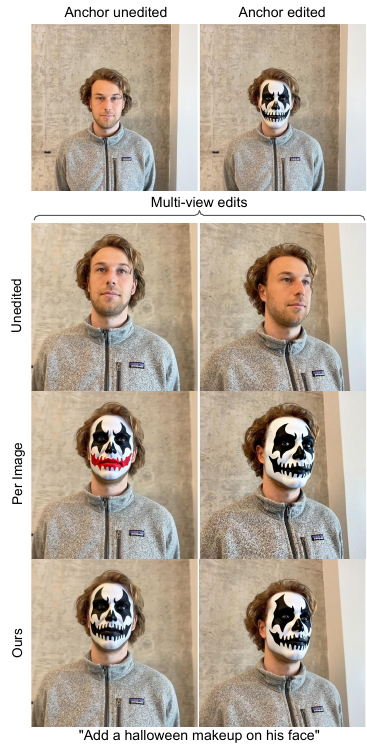}%
   \includegraphics[height=192.7px,trim={15px 0 0px 0},clip]{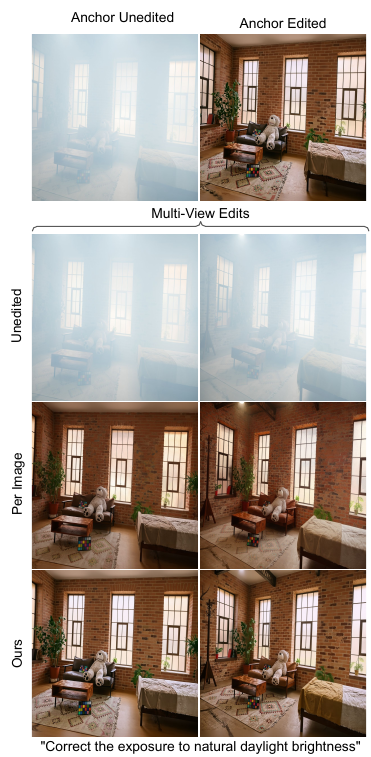}%
    \includegraphics[height=192.7px,trim={15px 0 0px 0},clip]{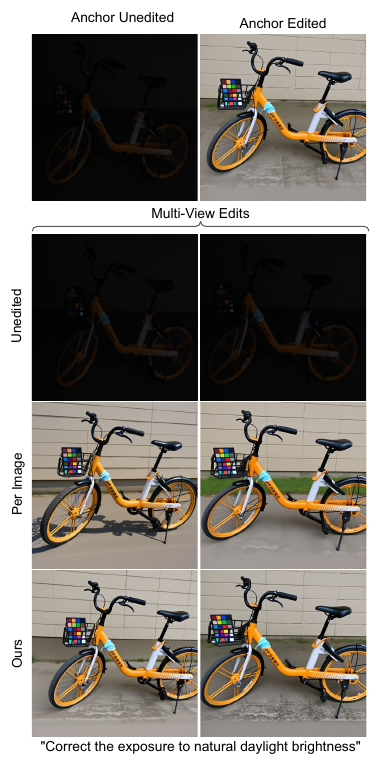}%
  \includegraphics[height=190px,trim={15px 0 0px 0},clip]{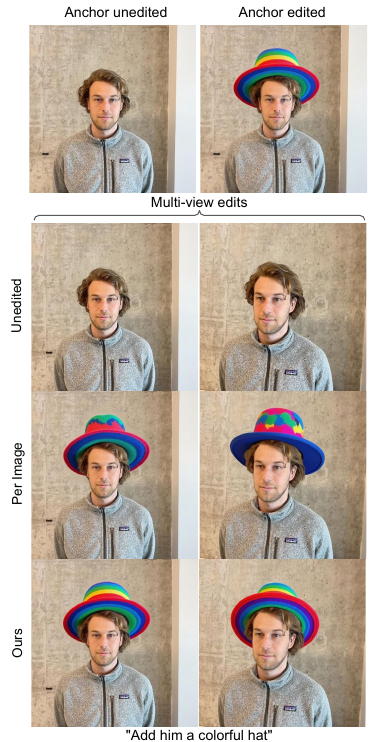}%
 
    \caption{Qualitative comparison using FLUX.1. The loss is computed between each image and an anchor image. In the three leftmost columns, we show rigid and near-rigid edits, and in the rightmost column, we show a non-rigid edit. The multi-view edits are more consistent with our approach, e.g., in the following details:
    (a) makeup paint colors, face pose, and gaze; (b) color and texture on the carpet and wall; (c) shadows and texture on the ground, and (d) hat colors.}
    \label{fig:flux_results}
\end{figure}
\begin{table}[t]
\resizebox{\textwidth}{!}{
\begin{tabular}{ll|ccccccccc}
\hline
\textbf{Ablation} &   & PSNR$\uparrow$ & SSIM$\uparrow$ & LPIPS$\downarrow$ & MEt3R$\downarrow$ & \multicolumn{1}{l}{CLIPdir$\uparrow$} & \multicolumn{1}{l}{CLIPsim$\uparrow$} & \multicolumn{1}{l}{CLIPimage$\uparrow$} & \makecell{Time (s/image)}\\ \hline
\multirow{3}{*}{\makecell[l]{Matched\\images}} 
 & 1 & 22.97           & 0.670           & 0.294              & 0.394              & 0.152                                  & 0.254                                  & 0.856                    & \textbf{28} \\
 & 2 & \textbf{23.15}  & \textbf{0.675}  & \textbf{0.290}     & \textbf{0.385}     & \textbf{0.153}                         & \textbf{0.255}                         & \textbf{0.857 }                  & 36 \\
 & 3 & 23.13           & 0.673           & 0.295              & 0.392              & 0.152                                  & 0.254                                  & 0.856                   & 44 \\ \hline
\multirow{3}{*}{\makecell[l]{Backward\\steps}}       
 & 0 & 21.71           & 0.654           & 0.294     & 0.396             & 0.164                      &       0.259              &     \textbf{0.868}           & \textbf{16} \\
 & 3 &23.15  & 0.675  &\textbf{ 0.290 }    & 0.385    & \textbf{0.153}                     & \textbf{0.255}                         & 0.857                    & 36\\
 & 6 & \textbf{23.51}  & \textbf{0.678}  & 0.304              &\textbf{ 0.382}              & 0.145                                  & 0.251                               &0.850                    & 57 \\ \hline
 \multirow{3}{*}{\makecell[l]{LPIPS patch \\ loss weight ($\lambda$)}}
 & 0 & 22.27  & 0.674 & 0.300            &0.403            & \textbf{0.165}                                &\textbf{0.260}                             &0.856                &\textbf{21}\\
 & 2 &\textbf{23.15}  & \textbf{0.675}  &\textbf{ 0.290 }    & \textbf{0.385}    & 0.153                     & 0.255                         &\textbf{ 0.857 }                   & 36\\
 & 5 & 22.56           & 0.674          & 0.313     &0.409           & 0.146   & 0.254  &   0.855          & 36\\ \hline
\end{tabular}
}
\vspace{1em}
\caption{Ablation of our method using InstructPix2Pix. We evaluate how many previously edited images to use for the consistency loss, the number of optimization steps for backward guidance, and the weight for the LPIPS patch loss.}
\label{tab:ablations}
\end{table}
It is a flow matching model for image generation and editing that demonstrates impressive quality and performs a wide range of edits, including image restorations such as removing smoke and adjusting exposure levels to achieve normal daylight brightness. It can also handle non-rigid edits such as object insertion or replacement (which InstructPix2Pix and pix2pix-Turbo cannot reliably do). It retains persistent character/object features through the edits. However, consistency in 3D pose, color, and texture remains limited, leading to inconsistencies across views. The improvements using our guidance are shown in Fig.~\ref{fig:flux_results}. We use the same optimization of the consistency loss in \eqref{eq:consistency_loss}, where for rigid and near-rigid edits (three leftmost  columns of Fig.~\ref{fig:flux_results}) we use matches between unedited images, and for non-rigid edits (rightmost column of Fig.~\ref{fig:flux_results}) we use matches between edited images. 
Note that our guidance works well for a diverse range of prompts.

\subsection{Ablation Studies}\label{sec:ablations}
We present ablation studies in Table \ref{tab:ablations} on important factors affecting performance and run-time. The ablations are provided using InstructPix2Pix \cite{brooks2022instructpix2pix}. We found that the number of images used to compute the matching loss (\ref{eq:consistency_loss}) saturated when using more than two images. We also found that using 3 backward steps in universal guidance provided the best trade-off between consistency and edit time, with 6 steps giving similar consistency metrics but at an increased computational cost. Additionally, the ablation shows that including the patch-based LPIPS loss gives improved consistency, while increasing its weight above $\lambda=2$ did not lead to additional improvements.

\section{Conclusion}
\label{sec:conculsion}
We present a flexible method for 3D consistent image editing. It guides the denoising process of a pre-trained single-image editing model by optimizing a correspondence-based objective so that matching points in different views are edited in a coherent way. We can improve the multi-view consistency for both rigid and near-rigid edits, as well as non-rigid edits, depending on which correspondences are used. Furthermore, we show that we can improve the multi-view consistency using a range of different image editing methods, namely diffusion models, one-step models, and flow matching models. We experimentally show that the consistency of the edited images is better compared to existing methods. We also demonstrate the possibility of editing a Gaussian splat model directly using both sparse and dense views.
\subsection*{Acknowledgments}
This work was supported by the Wallenberg AI, Autonomous Systems, and Software Program (WASP) funded by the Knut and Alice Wallenberg Foundation. Computational resources were provided by the National Academic Infrastructure for Supercomputing in Sweden (NAISS) at Chalmers Centre for Computational Science and Engineering (C3SE), partially funded by the Swedish Research Council under grant agreement no. 2022-06725, and by the Berzelius resource, provided by the Knut and Alice Wallenberg Foundation at the National Supercomputer
Centre.

% This document serves as an example submission to ECCV \ECCVyear{}.
% It illustrates the format authors must follow when submitting a paper. 
% At the same time, it gives details on various aspects of paper submission, including preservation of anonymity and how to deal with dual submissions.
% We advise authors to read this document carefully.

% The document is based on Springer LNCS instructions as well as on ECCV policies, as established over the years.

\bibliographystyle{splncs04}
\bibliography{main}

\appendix

\begin{center}
    \Large\textbf{Appendix}
\end{center}
\section*{Overview}
In this appendix we show additional qualitative results of our consistent multi-view image editing (\cref{sec:suppl_qualitative}), technical details related to both our multi-view editing (\cref{sec:suppl_mw_edit}) and the 3D Gaussian splat editing (\cref{sec:suppl_GS_edits}), and details of the dataset and prompts that were used (\cref{sec:suppl_scene_prompt_pairs}).

\textbf{We encourage the reader to watch video results on the project page \href{https://3d-consistent-editing.github.io/}{https://3d-consistent-editing.github.io/}} which presents a qualitative comparison across all applications. For full transparency and fair comparison, we also show the results from \emph{all} 21 scene-prompt pairs of the test set we use when evaluating the different methods. 

\section{Additional Qualitative Results}\label{sec:suppl_qualitative}

\begin{figure*}[]
    \centering
    \textbf{Edited Multi-View Images with InstructPix2Pix}
    \includegraphics[width=\textwidth,trim={0 22px 0 0},clip]{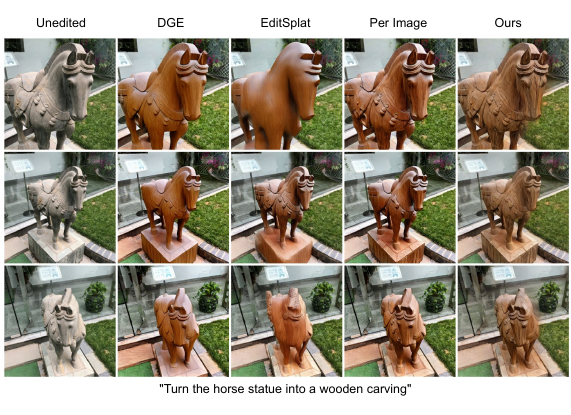}
    {\scriptsize \textsf{``Turn the horse statue into a wooden carving''}}\\[1em]
    \includegraphics[width=\textwidth,trim={0 22px 0 18px},clip]{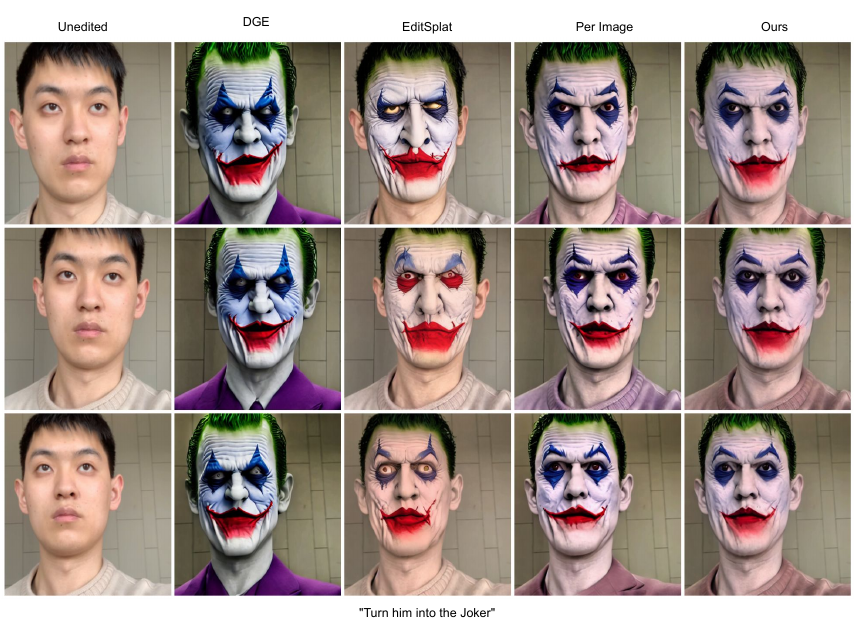}
    {\scriptsize \textsf{``Turn him into a Joker''}}\\
    \caption{Additional qualitative example of multi-view consistent image editing using methods all based on the image editing method InstructPix2Pix. We note that our method is able to preserve details better than the other methods, as seen, e.g., on the saddle and the area around the eyes.}
    \label{fig:image_consistency_sup_wooden_joker}
\end{figure*}

\begin{figure*}[]
    \centering
    \textbf{Edited Multi-View Images with pix2pix-Turbo}
    \includegraphics[width=\textwidth,trim={0 5px 0 0},clip]{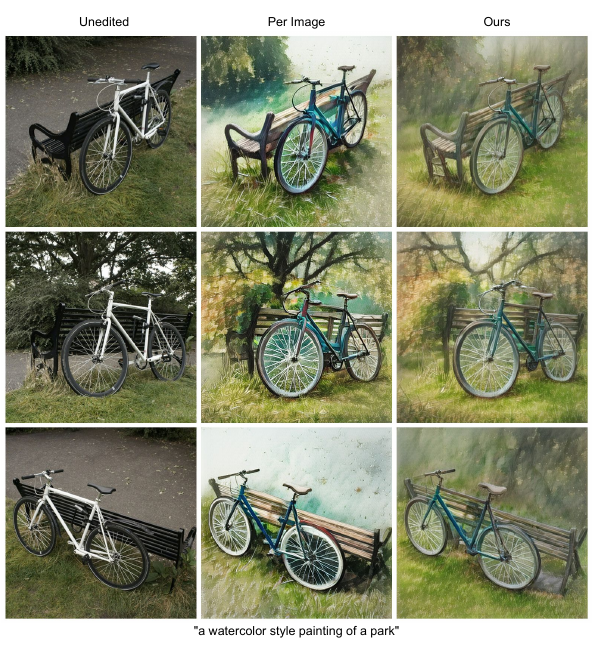}
    \caption{Additional qualitative example of multi-view consistent image editing with pix2pix-Turbo. We note that our method gives a more consistent overall color, e.g., on the bicycle frame, and that the background is more consistent.}
    \label{fig:image_consistency_sup_watercolor}
\end{figure*}

\begin{figure*}[]
    \centering
    \textbf{Edited Multi-View Images with pix2pix-Turbo}
    \includegraphics[width=\textwidth,trim={0 5px 0 0},clip]{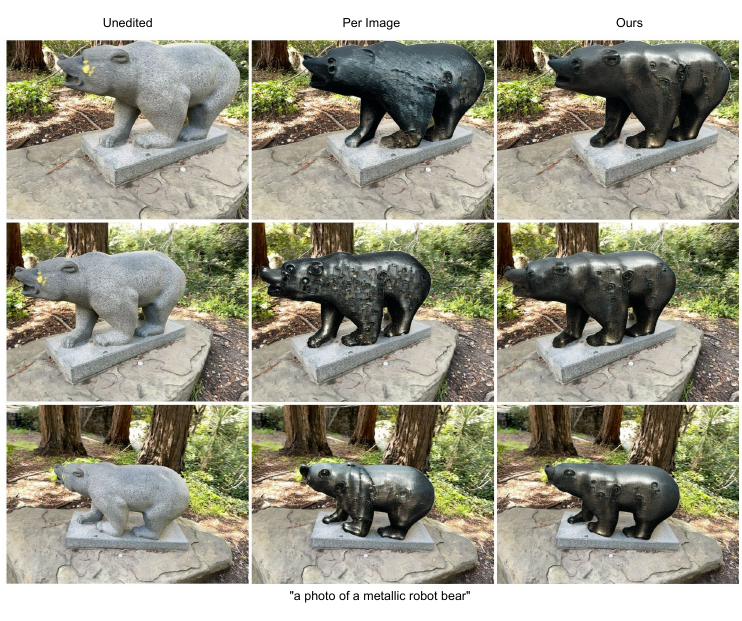}
    \caption{Additional qualitative example of multi-view consistent image editing with pix2pix-Turbo. We note that with our method the texture on the bear is consistent across the views, and that realistic lighting reflections are captured.}
    \label{fig:image_consistency_sup_robot}
\end{figure*}

\begin{figure}[]
    \centering
    \textbf{Edited Multi-View Images with FLUX.1}
    \includegraphics[height=185px,trim={0 0 0 0},clip]{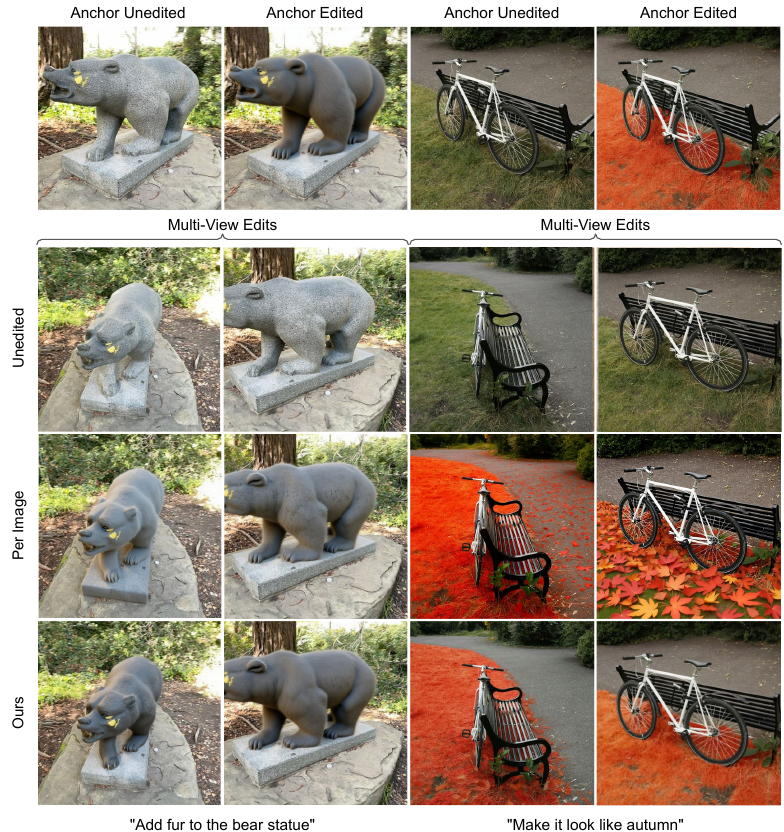}
  \includegraphics[height=185px,trim={0px 0 0px 0},clip]{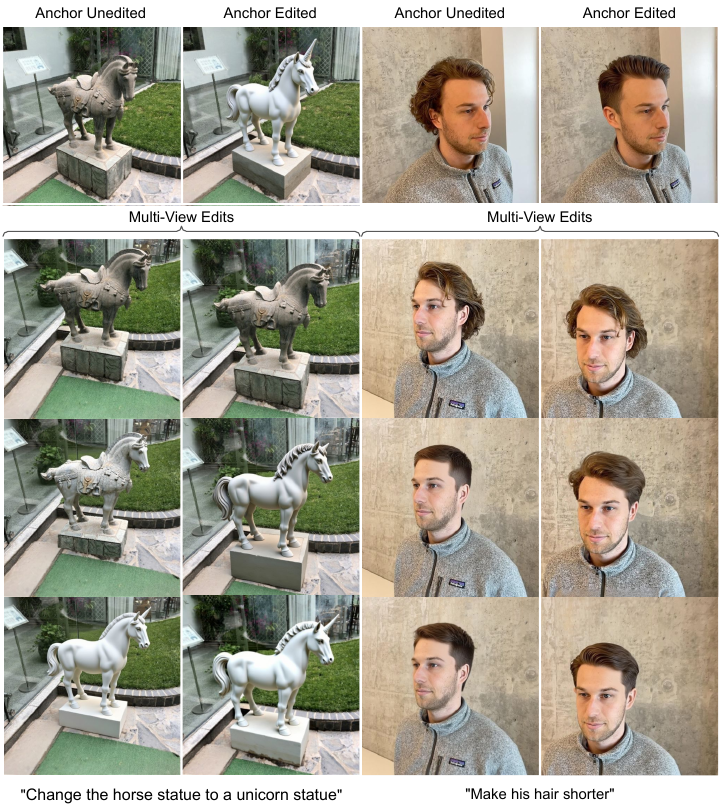}
    \caption{Additional qualitative comparison using FLUX.1. The loss is computed between each image and an anchor image. On the left we show rigid and near-rigid edits, and on the right we show non-rigid edits. The multi-view edits are more consistent with our approach, e.g., in the following details: (a) fur color and pattern, texture of the platform; (b) appearance of grass and pavement; (c) texture and shape of the statue, and (d) hair length and overall style.}
    \label{fig:sup_flux_results}
\end{figure}

We provide multiple additional examples of our multi-view consistent image editing in this section.
\paragraph{Multi-View Image Editing} 
We show results for InstructPix2Pix \cite{brooks2022instructpix2pix} in Fig. \ref{fig:image_consistency_sup_wooden_joker}, where we note that, e.g., the eyes and the surrounding regions are more consistent for our method and the details on the saddle are better preserved across views when using our method. For pix2pix-Turbo \cite{parmar2024one}, we see in Fig. \ref{fig:image_consistency_sup_watercolor} that the overall colors and backgrounds are more consistent for our method and in Fig. \ref{fig:image_consistency_sup_robot} we note that the texture on the bear is more consistent with our editing. Finally, we show additional results for FLUX.1 \cite{flux} in Fig. \ref{fig:sup_flux_results}, where it can be seen that our method improves color and texture consistency in the fur of the bear and the grass in front of the bicycle. We also observe consistent shape and appearance for the unicorn statue and improved geometry for the changed hairstyle.
\paragraph{3DGS Editing}
On the project page we provide additional video results showing Gaussian splat renderings for both the dense and sparse view setups. Additionally, we show our method and the methods we compare to for all 21 scene-prompts pairs in our test set, using both InstructPix2Pix \cite{brooks2022instructpix2pix} and pix2pix-Turbo \cite{parmar2024one}.

\begin{figure*}[h]
    \centering
    \includegraphics[width=\linewidth]{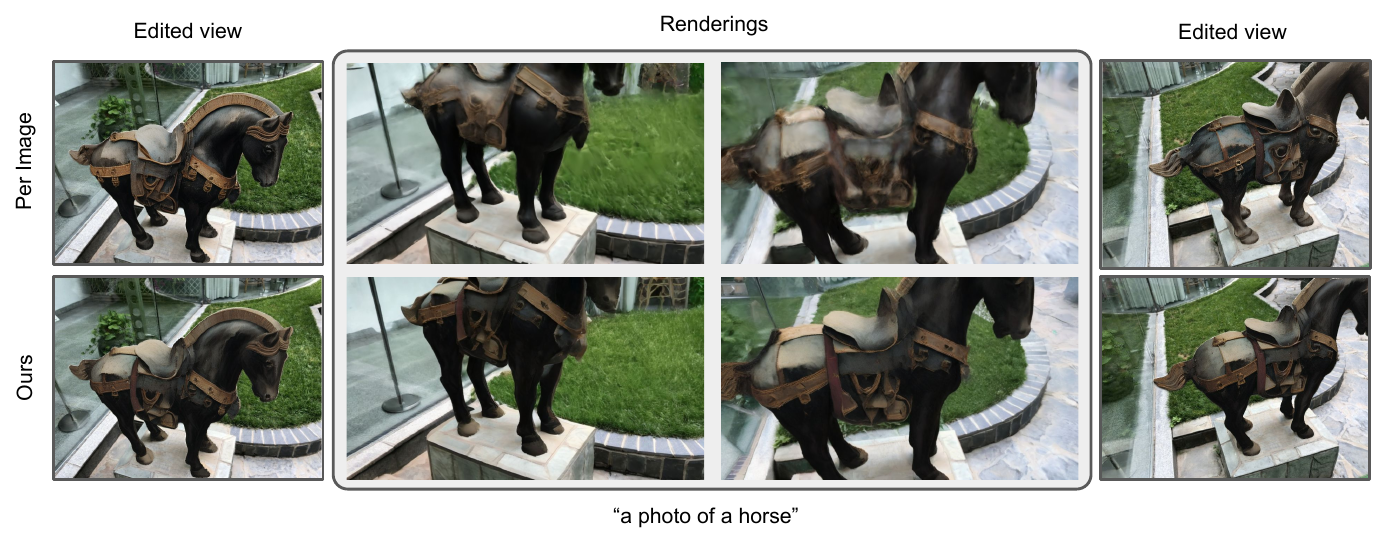}
    \caption{Additional example of editing a few sparse views and using ViewCrafter to interpolate. We note that the edited views are more consistent for our method compared to per image edits, which can be seen, e.g., in the saddlebag in both the edited images and the Gaussian splat renderings.}
    \label{fig:sup_sparse}
\end{figure*}

\paragraph{Sparse 3DGS Editing}
We show an additional example of the sparse editing in Fig. \ref{fig:sup_sparse}, where we see that the renderings of the Gaussian splat model are more consistent when using our edited images compared to the per image edits.

\section{Multi-View Image Editing}\label{sec:suppl_mw_edit}

\paragraph{Image Editing Methods}
We test our method together with two different diffusion based image editing methods, InstructPix2Pix \cite{brooks2022instructpix2pix} and the single-step model pix2pix-Turbo \cite{parmar2024one}, as well as the flow-matching based method FLUX.1 \cite{flux}. For \textit{InstructPix2Pix} we use $20$ denoising steps and guidance scales $s_T=7.5$ and $s_I=1.5$, which are the same choices as in DGE and EditSplat. The single-step model \textit{pix2pix-Turbo} is trained to take a Canny edge map and generate an image based on this and a given text prompt. So the editing process with pix2pix-Turbo is to first convert the image to a Canny edge map that is then used to generate an edited image, which leads to the appearance of the whole image being changed even if a local edit is desired. To address this we utilize a segmentation mask to ensure that only the desired object is changed and that the rest of the image remains unchanged. For the flow-matching based model FLUX.1 \cite{flux} we use $28$ denoising steps. We also take advantage of the fact that denoising trajectories for flow-matching based models are approximately linear, making it possible to avoid computing gradients through all the denoising steps, as described in \cite{SONIC}. This assumption of linear denoising trajectories should in theory be a perfect approximation \cite{lipman2023flow,liu2023flow}, but in practice an approximation error still exists. We find that, while using this approximation, we are still able to stably converge to initial seeds that improve consistency.

\paragraph{Image Matching Details}
We use the robust dense matcher RoMa \cite{edstedt2024roma}, which is a robust dense matcher that also returns certainties. We only use matches with a certainty over $0.05$ and include a maximum of $50~000$ matches. For the LPIPS patch loss we use the perceptual loss \cite{zhang2018perceptual} applied to patches centered around the matches. We randomly select $1~500$ correspondences out of the current matches and extract patches of size $64 \times 64$ centered at these positions. 

\begin{figure*}[ht!]
    \centering
    \includegraphics[width=\linewidth]{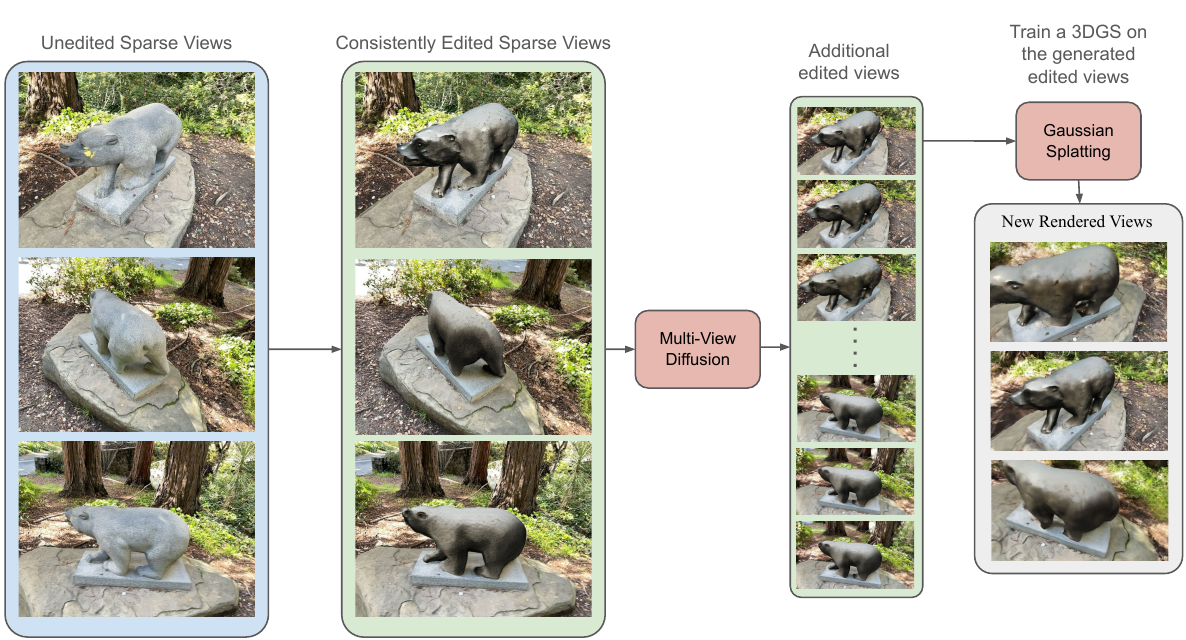}
    \caption{Overview of the sparse editing pipeline, that from a few sparse views generates a Gaussian splat model of the edited scene. Our consistent editing method is used to edit the given sparse views and a Multi-View Diffusion network is then used to generate additional edited images of the same scene. A Gaussian splat model can then be trained on all these views, and we can render novel views using the Gaussian splat model.}
    \label{fig:system_figure_sparse}
\end{figure*}

\paragraph{Sparse Editing}
We show an overview of the sparse editing in Fig. \ref{fig:system_figure_sparse}. For the sparse editing we edit 3-4 images and then utilize the multi-view diffusion model ViewCrafter \cite{yu2024viewcrafter} to interpolate between these views to generate 50-75 additional views of the scene. These additional edited views can then be used to train a Gaussian model representing the edited scene. In this setup there is no existing Gaussian model of the unedited scene available since we only have a few images of the scene. We thus train a Gaussian model from scratch based on the generated edited views, using $5~000$ iterations. Training from scratch instead of editing a Gaussian splat model makes this setup more sensitive to inconsistencies in the edited views. Since a new Gaussian model needs to be generated from the edited images, and we do not have any prior geometry from the Gaussian splat model of the unedited images, inconsistencies in the images can easily lead to incorrect geometry or strong view-dependent effects. Another reason is that inconsistencies in the edited sparse views can lead to significant appearance change in the additional views generated by ViewCrafter, which again lead to difficulties of training a Gaussian splat model.

\section{Gaussian Splat Editing}\label{sec:suppl_GS_edits}

In this section, we describe the differences in the training of the edited Gaussian splat models between our method and the methods we compare to. When we test EditSplat \cite{lee2025editsplat} and DGE \cite{chen2024dge}, we use their provided code without any changes. We refer to these papers for full details and provide a brief overview here.

\textbf{Ours.} We use the standard Gaussian splatting training procedure from the original paper \cite{kerbl20233d}, except that we limit the training to 20 epochs (800-2500 iterations depending on the number of images in the scene), since we resume the training from a Gaussian splat model of the unedited images. Different from the original Gaussian splat training settings, we used a loss function where the L1 and LPIPS rendering losses have equal weights, similar to earlier editing methods \cite{lee2025editsplat,haque2023instruct}. 

\textbf{EditSplat.} Similarly to ours, EditSplat also first edits the images and then uses those images to update the Gaussians. There are several techniques used to edit the Gaussians. Based on the prompt and edit model, they compute attention weights over the images, indicating regions where the prompt has a large influence and where the Gaussians should change. The Gaussians are trimmed so that a fixed fraction of the Gaussians with large attention weights are excluded from the editing. The motivation is that retraining excessive source attributes, as indicated by the regions with large attention weights, is detrimental for the optimization to the edited images.

\textbf{DGE.} DGE also initially performs a multi-view consistent editing process that is used to update an existing Gaussian model of the unedited images. But instead of just performing one update step, they perform an iterative refinement where they render images from the updated Gaussians and repeat the editing process, re-updating the 3D model. This process is repeated for a total of 3 iterations, which was the default value in their released code.

\section{Scene-Prompt Pairs}\label{sec:suppl_scene_prompt_pairs}

We show all scene-prompt pairs used in the test set in Table \ref{tab:scene_prompt_pairs}. Most of the pairs are the same as used in prior work \cite{lee2025editsplat}. We additionally show all pairs in the validation set in Table \ref{tab:val_scene_prompt_pairs}. Note that different scenes are used in the validation and test sets.

\begin{table}[h]
\centering
\scriptsize
\begin{tabularx}{\columnwidth}{l Y Y Y}
\textbf{Scene} & \textbf{Source Description} & \textbf{Target Description} & \textbf{Editing Instruction} \\ \hline

% Face
Face & a photo of a face of a man
     & a photo of a face of a clown
     & Make his face look like a clown \\

Face & a photo of a face of a man
     & a photo of a marble sculpture
     & Make his face resemble that of a marble sculpture \\

Face & a photo of a face of a man
     & a Vincent Van Gogh painting
     & Make him look like a Vincent Van Gogh painting \\ \hline

% Fangzhou
Fangzhou & a photo of a face of a man
         & a photo of the face of the Joker
         & Turn him into the Joker \\

Fangzhou & a photo of a face of a man
         & a photo of the face of Steve Jobs
         & Turn him into Steve Jobs \\

Fangzhou & a photo of a face of a man
         & a photo of a face of a man with Maasai face paint
         & Give him Maasai face paint \\ \hline

% Garden
Garden & a photo of an outdoor garden
       & a photo of a foggy outdoor garden
       & Make it foggy \\

Garden & a photo of an outdoor garden
       & a photo of a snowy outdoor garden
       & Make it snowy \\ \hline

% Bicycle
Bicycle & a photo of a park
        & a photo of a Namibian desert
        & Turn the ground into a Namibian desert \\

Bicycle & a photo of a park
        & a watercolor style painting of a park
        & Make the entire scene look as if it’s painted in watercolor style \\ \hline

% Bear
Bear & a photo of a bear statue
     & a photo of a metallic robot bear
     & Turn the bear statue into a metallic robot \\

Bear & a photo of a bear statue
     & a photo of a panda
     & Turn the bear statue into a panda \\ \hline

% Person
Person & a photo of a person
       & a photo of a person in Minecraft
       & Turn him into a Minecraft character \\

Person & a photo of a person
       & a photo of a person wearing clothes with a pineapple pattern
       & Make the person wear clothes with a pineapple pattern \\

Person & a photo of a person
       & a photo of a person wearing a suit
       & Make the person wear a suit \\ \hline

% Bonsai
Bonsai & a photo of a bonsai
       & a photo of a snowy bonsai
       & Make the bonsai snowy \\

Bonsai & a photo of a bonsai
       & a photo of a bonsai with yellow petals
       & Make the bonsai have yellow petals \\

Bonsai & a photo of a bonsai
       & a photo of a bonsai made of paper
       & Change the bonsai to look like it’s made of paper, folded into intricate origami shapes \\ \hline

% Stone Horse
Stone Horse & a photo of a horse statue
            & a photo of a horse made of wood
            & Turn the horse statue into a wooden carving \\

Stone Horse & a photo of a horse statue
            & a photo of a horse made of jade
            & Turn the stone horse into a jade carving \\

Stone Horse & a photo of a horse statue
            & a photo of a zebra
            & Make the stone horse a zebra \\

\end{tabularx}

\caption{All 21 scene-prompt pairs used as a test set for evaluation.}
\label{tab:scene_prompt_pairs}
\end{table}

\begin{table*}[h]
\centering
\small
\begin{tabularx}{\textwidth}{l Y Y Y}
\textbf{Scene} & \textbf{Source Description} & \textbf{Target Description} & \textbf{Editing Instruction} \\ \hline
\scriptsize
% Kitchen
Kitchen & a photo of a lego excavator
        & a Vincent Van Gogh painting of a lego excavator
        & Turn it into a Vincent Van Gogh painting \\ \hline

% MaleFace
MaleFace & a photo of a face of a man
         & a photo of a face of a man with a bandana
         & Give him a bandana \\

MaleFace & a photo of a face of a man
         & a photo of a face of a very old man
         & Make him look very old \\ \hline

% Campsite
Campsite & a photo of a campsite
         & a photo of a campsite in the Sahara desert
         & Make the ground look like the Sahara desert \\

Campsite & a photo of a campsite
         & a photo of a campsite with snow on the ground
         & Make the ground snowy \\ \hline

% Vasedeck
Vasedeck & a photo of a vase with flowers
         & a photo of a vase with some red flowers
         & Make some of the flowers red \\

Vasedeck & a photo of a vase with flowers
         & a photo of a vase with yellow and blue flowers
         & Make the flowers yellow and blue \\

\end{tabularx}

\caption{All 7 scene–prompt pairs used as a validation set when choosing hyperparameters and performing ablation studies for our method.}
\label{tab:val_scene_prompt_pairs}
\end{table*}

\end{document}